\documentclass[lettersize,journal]{IEEEtran}
\usepackage{amsfonts}
\usepackage{array}
\usepackage[caption=false,font=normalsize,labelfont=sf,textfont=sf]{subfig}
\usepackage{textcomp}
\usepackage{stfloats}
\usepackage{url}
\usepackage{verbatim}
\usepackage{graphicx}
\usepackage{cite}

\usepackage{amssymb}
\usepackage{multirow}
\usepackage{pifont} 
\usepackage{utfsym} 
\usepackage{algpseudocode}

\usepackage{algorithm}  
\usepackage{algpseudocode}  
\usepackage{amsmath}

\usepackage{physics}

\begin{document}

\title{DivDiff: A Conditional Diffusion Model for Diverse Human Motion Prediction
}

\author{Hua Yu, Yaqing Hou, Wenbin Pei, Yew-Soon Ong, \emph{Fellow, IEEE}, and Qiang Zhang
\thanks{Manuscript created xxx. 
This work was supported in part by the National Key Research and Development Program of China (No. 2021ZD0112400), the National Natural Science Foundation of China under Grant 62372081, the Young Elite Scientists Sponsorship Program by CAST under Grant 2022QNRC001, the NSFC-Liaoning Province United Foundation under Grant U1908214, the 111 Project, No.D23006, and the Fundamental Research Funds for the Central Universities under grant DUT21TD107, DUT22ZD214. (Corresponding authors: Qiang Zhang)

Hua Yu, Yaqing Hou, Wenbin Pei and Qiang Zhang are with the School of Computer Science and Technology, Dalian University of Technology, Dalian 116622, China (e-mail: yhiccd@mail.dlut.edu.cn;
houyq@dlut.edu.cn; peiwenbin@dlut.edu.cn; zhangq@dlut.edu.cn)

Yew-Soon Ong is with the School of Computer Science and Engineering, Nanyang Technological University, Singapore 639798, Singapore (e-mail: asysong@ntu.edu.sg).
}}

\markboth{IEEE Transactions on Multimedia}%
{Shell \MakeLowercase{\textit{et al.}}: A Sample Article Using IEEEtran.cls for IEEE Journals}


\maketitle

\begin{abstract}
Diverse human motion prediction (HMP) aims to predict multiple plausible future motions given an observed human motion sequence.
It is a challenging task due to the diversity of potential human motions while ensuring an accurate description of future human motions.
Current solutions are either low-diversity or limited in expressiveness.
Recent denoising diffusion models (DDPM) hold potential generative
capabilities in generative tasks. However, introducing DDPM directly into diverse HMP incurs some issues. Although DDPM can increase the diversity of the potential patterns of human motions, the predicted human motions become implausible over time because of the significant noise disturbances in the forward process of DDPM. This phenomenon leads to the predicted human motions being hard to control, seriously impacting the quality of predicted motions and restricting their practical applicability in real-world scenarios. 
To alleviate this, we propose a novel conditional diffusion-based generative model, called DivDiff, to predict more diverse and realistic human motions. Specifically, the DivDiff employs DDPM as our backbone and incorporates Discrete Cosine Transform (DCT) and transformer mechanisms to encode the observed human motion sequence as a condition to instruct the reverse process of DDPM.
More importantly, we design a diversified reinforcement sampling function (DRSF) to enforce human skeletal constraints on the predicted human motions. DRSF utilizes the acquired information from human skeletal as prior knowledge, thereby reducing significant disturbances introduced during the forward process. Extensive results received in the experiments on two widely-used datasets (Human3.6M and HumanEva-I) demonstrate that our model obtains competitive performance on both diversity and accuracy.
\end{abstract}


\section{Introduction}
Human motion prediction has gained tremendous attention in the realm of human-robot interactions. This field has wide-ranging applications, such as human motion understanding \cite{9795092,9667292}, autonomous driving \cite{9772775,9580659}, and animation \cite{hua2022towards,kim2021motion,zhang2021spatially}. 
Prior works for human motion prediction mainly focus on deterministic methods \cite{aksan2021spatio,cui2020learning}. Given the observed human motion sequence, the designed model aims to predict the most likely future human motions, such as Recurrent Neural Networks (RNNs)-based \cite{Fragkiadaki_2015_ICCV,hou2021local}, Graph Convolutional Networks (GCNs)-based \cite{ma2022multi,ding2022towards}, Transformer-based \cite{aksan2021spatio,hua2022towards} methods.
However, human motions are often stochastic owing to the complex environments and the uncertainty of human intentions.
The deterministic methods are inappropriate for real-world applications in a more complex environment, ignoring the ``one-to-many'' nature of the problem.
For example, the observed human motion is ``walking'', the future human motions might be ``jumping'', ``walking'' or ``running''. It means that our designed model is anticipated to predict potential motions corresponding to various walking patterns. 
Under these settings, the designed method is supposed to learn the multi-modal distribution for diverse human motions. 

Given a historical motion sequence, the designed stochastic method is supposed to predict multiple possible future motions, i.e., to fully capture all potential patterns of human motions. However, this task is still complex due to the wide variety of human motions and intricate interactions with the surroundings, making it challenging to model dynamic human motions \cite{aliakbarian2020stochastic,gu2022stochastic}.
Deep generative models, such as variational auto-encoders (VAEs) \cite{cai2021unified,mao2021generating,mccarthy2020addressing} and generative adversarial networks (GANs) \cite{gu2022stochastic,gui2018adversarial,ma2022multi}, have demonstrated their effectiveness in capturing the data distribution cross all possible human motions. 
For the VAE-based methods, Yan et al. \cite{yan2018mt} proposed motion transformation variational auto-encoders (MT-VAE) for generating multiple plausible motions through a stochastic sampling of the feature transformations. 
Sadegh et al. \cite{aliakbarian2021contextually} incorporated a latent variable sampling into the representation of the observed human motion sequences, which can serve as a source of diversity. 
Yuan et al. \cite{yuan2020dlow} investigated a trainable sampling strategy called DLOW, enhanced with prior knowledge to promote the diversity of future predictions, aiming to reduce redundancy.
Although significant progress has been made, the VAE-based methods usually tend to capture the major patterns of human motions, i.e., the most likely human motions, while ignoring the minor patterns, i.e., low-probability motions.

For GAN-based methods, Barsoum et al. \cite{barsoum2018hp} proposed a probabilistic generative adversarial network (HP-GAN) along with a noise variable to learn a probability density function of future human motions. This method utilizes a random vector $\textbf{\emph{z}}$ from a prior distribution to control the generation of multiple possible future human poses. However, their method tends to ignore the random vectors. To alleviate this issue, Kundu et al. \cite{kundu2019bihmp} introduced a novel bidirectional method (BiHMP-GAN) for human motion prediction. Specifically, they design a discriminator to regress a sampled extrinsic factor $r$, which is used for the prediction of the corresponding future dynamics. 
Nevertheless, the GAN-based methods usually neglect the influence of the variations, and the training process is often unstable due to complex adversarial learning process \cite{aliakbarian2020stochastic,bowman2015generating}. 

\begin{figure*}
    \centering 
    \includegraphics[width=6.6in]{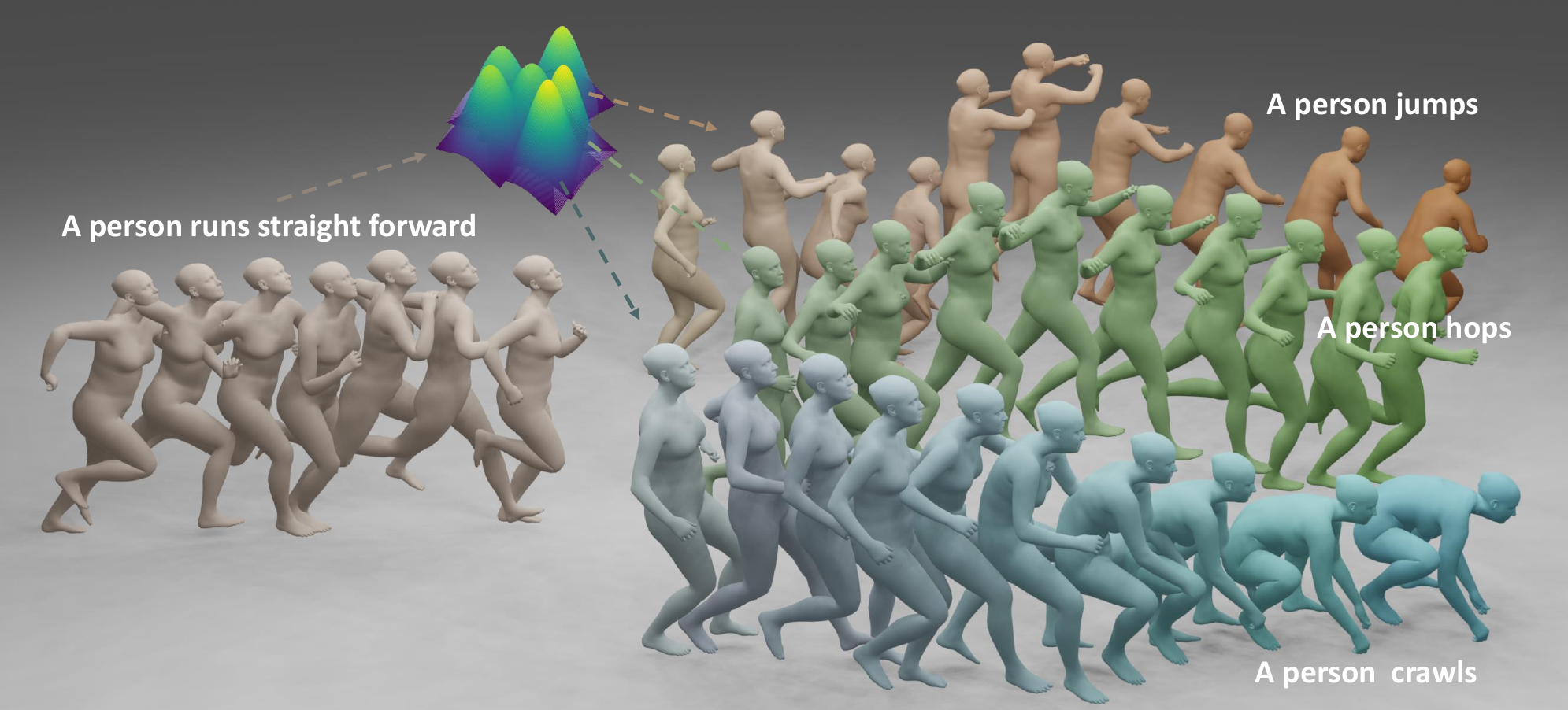}
    \caption{The proposed DivDiff method aims to predict the future stochastic human motions (right) given the past human sequence (left). For example, the observed human motion is ``a person runs straight forward", the potential future human motions might be ``a person jumps", ``a person hops'', ``a person crawls" and so on.
    The extensive experiments demonstrate that the proposed DivDiff has significantly enhanced the diversity and fidelity of the predicted human motions.}
 \label{fig:1}
\end{figure*}

Motivated by the recent advances in synthesis and generation tasks \cite{wei2022human,yuan2022physdiff}, Denoising Diffusion Probabilistic Model (DDPM) \cite{ho2020denoising} is utilized to predict future human motions \cite{barquero2022belfusion}.
The forward process of DDPM gradually adds Gaussian noise to the input sequences controlled by the variance schedule $\beta$ until the whole sequence is completely corrupted as Gaussian noise. The reverse process of DDPM aims to recover the realistic human motion from the noise.
After conducting extensive experiments, we have noticed that introducing DDPM into the diverse human motion prediction may incur some issues.
When using a smaller number of denoising steps (typically $\leq$50) in the reverse diffusion process, the predicted human motions might be more diverse but usually lead to low accuracy. On the other hand, when using a larger number of denoising steps (typically \textgreater500), though the accuracy of the results is improved, the diversity of predicted human motions is declined. This is mainly because using more diffusion steps usually generates realistic human motions and is not capable of learning multiple patterns of motions, while too few steps cannot generate realistic human motions due to significant disturbances. In addition, more denoising steps lead to a slow inference process.
The noise parameter in the late stage of the forward process fluctuates significantly compared to the early stage, causing noticeable disturbances in the late stage of the reverse process. 
The aforementioned phenomenon involves a delicate balance between the diversity and accuracy of predicted human motions. However, this issue has received limited attention in existing works on human motion prediction. In view of this, it is prompted to investigate an effective method that promotes high diversity and motion quality.

To facilitate this, we propose a novel method called DivDiff, which is based on a conditional DDPM, for more diverse and realistic 3D human motion prediction. 
Specifically, for the forward process of DDPM, the motion sequence is diffused into white noise by progressively adding Gaussian noise at each time step.
For the reverse process of DDPM, the proposed method designs a state embedding to instruct the reverse diffusion process, which consists of encoding the observed sequence using Discrete Cosine Transform (DCT) \cite{mao2019learning} and transformers \cite{vaswani2017attention}. The aim is to capture temporal smooth and spatial relations between human joints.
In addition, a novel Diversified Reinforcement Sampling Function (DRSF) is introduced into the late stage of the reverse process due to the significant disturbances of added noise in the forward process.
DRSF leverages graph convolutional network \cite{cui2020learning} to obtain the human skeleton constraint information, which functions as prior knowledge to effectively capture the internal relationship between human bones. Thereby improving the quality of predicting human motions.
This operation enables the prediction of realistic and diverse human motions while accelerating the inference process with fewer denoising steps. 
As shown in Fig. 1, given an observed human sequence, the proposed DivDiff method is capable of predicting multiple and realistic human motions.

Overall, the contributions of this work are summarized as follows:

\begin{itemize}
\item{We propose an efficient diffusion-based generative model, called DivDiff, for 3D diverse human motion prediction. Our method is built upon the conditional Denoising Diffusion Probabilistic Model and is capable of predicting more stochastic and realistic human motions. }
\item{We introduce a novel diversified reinforcement sampling strategy (i.e., DRSF) and integrate it into a latent diffusion model to learn diversified Gaussian distributions, thereby promoting the diversity of predicted human motions while assuring accuracy.}
\item{The proposed method is evaluated on two widely-used benchmarks (i.e., Human3.6M and HumanEva-I) in the comprehensive experiments. The obtained results demonstrate the effectiveness of the proposed method over the state-of-the-art approaches for 3D human motion prediction tasks.}
\end{itemize}


\begin{figure*}
    \centering 
    \includegraphics[width=7.3in]{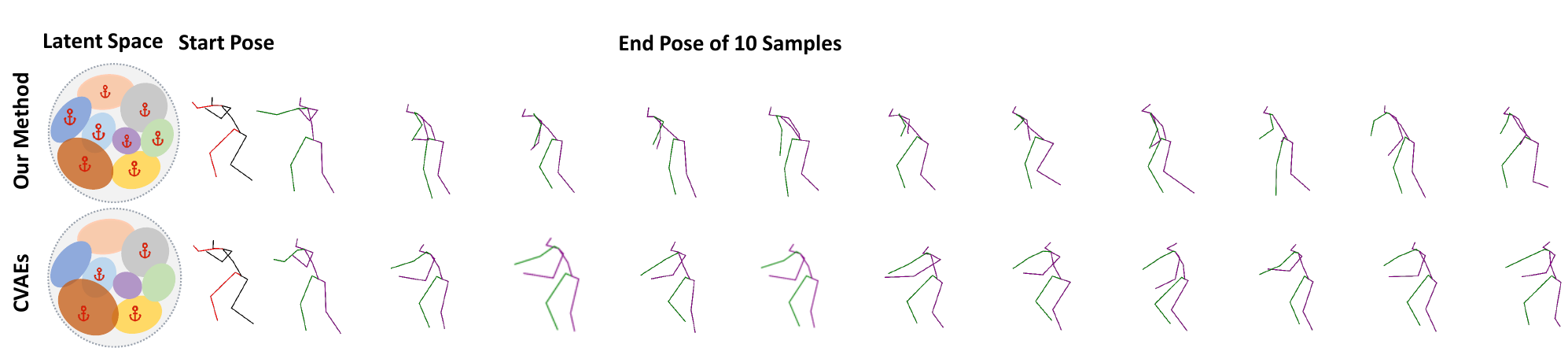}
    \caption{The generated samples from the proposed method are capable of covering more patterns (colored ellipses) compared to the CVAEs method. In the feature space, our method is able to capture a diverse set of future human motion patterns. However, the CVAEs method generates a large number of samples that are mainly concentrated on the major patterns of the motion distribution, failing to encompass the minor patterns.} 
 \label{fig:2}
\end{figure*}

\section{Related Work}
\subsection{Deterministic Human Motion Prediction}
Most previous efforts on human motion prediction have focused on predicting the most likely future human motions, and tend to utilize deterministic models \cite{aksan2021spatio,li2020dynamic}. Specifically, these methods cast motion prediction as a regression task where one output is produced given the observed sequence. Considering the efficacy of recurrent neural networks (RNNs) \cite{Fragkiadaki_2015_ICCV} in capturing temporal dependencies in time-series data, numerous researchers \cite{hou2021local,men2020quadruple,wang2019combining} resort to employing RNNs-based variant for this task.
Nevertheless, these approaches frequently encounter issues of discontinuity and error accumulation due to their structure characteristics.

To mitigate this phenomenon, Graph Convolutional Networks (GCNs) \cite{dang2021msr,ding2022towards,li2020dynamic,mao2019learning} based methods have been leveraged as a solution for predicting future human motions, as they are effective in capturing the intricate spatial and temporal relations between human motions. Nevertheless, GCNs have some limitations in modeling long-range dependencies, because the convolutional operations have a limited receptive field. To address this issue, the transformer mechanism \cite{vaswani2017attention} has been introduced as an alternative to handle the human motion prediction \cite{aksan2021spatio,hua2022towards}. The transformer mechanism utilizes the self-attention mechanism to model the dependencies between any two positions in the sequence, regardless of their distance. This makes it well-suited for capturing long-range dependencies. 

While remarkable progress has been made, these methods seldom consider the inherent uncertainty of human motions under the complex real world. This refers to the fact that there are diverse solutions for human motion prediction. Different from these methods, this work attempts to predict multiple possible human motions given a single observed human sequence.

\begin{figure*}[htb]
 \centering 
 \includegraphics[width=6.9in]{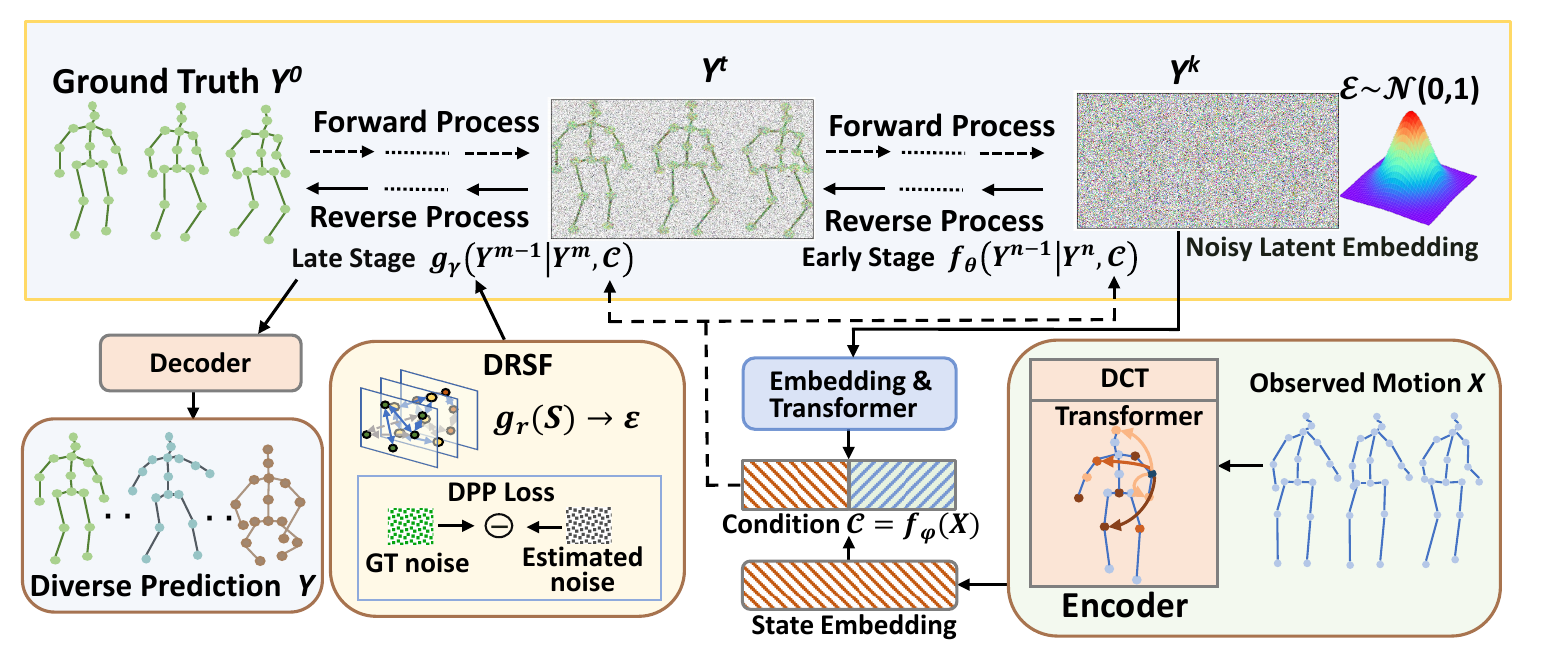}
 \caption{The illustration of the proposed DivDiff method. The observed sequence $X$ is encoded by the DCT and transformer as a state embedding, which serves as a condition to guide the diffusion process. The ground truth future motions $\mathbf{\emph{Y}}^0$ incorporates $K$ times noise variables to the whiten noise $\mathbf{\emph{Y}}^K$. In the late stage of the reverse process, the designed DRSF utilizes GCN to learn the relationships between human skeletons and serves as prior knowledge. Meanwhile, DPP loss is employed to predict random noise and control the quality of the predicted results.}
 \label{fig:2}
\end{figure*}

\subsection{Diverse Human Motion Prediction}
Given the inherent indeterminacy of future human motions,
the generative models like variational autoencoders (VAEs) \cite{cai2021unified,mao2021generating}, generative adversarial networks (GANs) \cite{gui2018adversarial,kocabas2020vibe,odena2017conditional} and denoising diffusion probabilistic model
(DDPM) \cite{ho2020denoising,barquero2022belfusion} are appropriate methods for diverse human motion prediction due to their capability of generating a diverse set of valid solutions.
Generally, VAEs \cite{mao2021generating,walker2017pose} utilize the encoder-decoder mechanism to handle the task by capturing the probability distribution of human motions. 
In these methods, the major modes (with higher likelihood) corresponding to a particular human motion pattern will more likely generate samples, minor patterns (those with lower likelihood) will almost not generate any samples. For example, as shown in Fig. 2, 
the VAEs method generates a large number of samples, which are mainly concentrated on the major patterns of the data distribution and fail to cover the minor patterns. Our method is capable of capturing the potential likely patterns of human motions.
GANs \cite{gui2018adversarial,kocabas2020vibe} employ a generator to produce multiple future human motions, while a discriminator ensures that the generated motions are similar to the ground truth.
However, these methods are often prone to mode collapse and involve complex adversarial learning. 
This phenomenon may result in the predicted motions being limited to a narrow set of possibilities, instead of capturing the full diversity of the observed human motions.

More recently, DDPM \cite{ho2020denoising} has been proposed as an alternative generative approach for the task of human motion generation. For example, Tevet et al. \cite{barquero2022belfusion} designed a Motion Diffusion Model (MDM) for various human motion generation, enabling different modes of conditioning, and different generation tasks. Ye et al. \cite{yuan2022physdiff} have attempted to alleviate this issue by introducing physics simulation, which is based on a pre-trained physical model through reinforcement learning. However, their approach is computationally intensive and the physics simulator cannot be trained together with the proposed method. 
Different from human motion generation task, when applying DDPM to human motion prediction, there exists a trade-off between the diversity and accuracy of the predicted human motions. 
It should be noted that the “diversity” in motion prediction task differs greatly from generation task. The generation task aims to generate different sequences under the same semantic meaning, while the motion prediction task aims to predict various sequences to capture the potential motion patterns.
It is significant to explore an efficient method for diverse human motion prediction while assuring the quality of the results.



\section{The Proposed Method}
In this section, we first briefly introduce the problem formulation. Then, the preliminaries are described for the mechanism of acquiring spatio-temporal information of human motions. Subsequently, the forward motion diffusion process is reported for diverse human motion prediction. Finally, the diversity reinforcement sampling function is presented in the conditional reverse process to enhance the diversity and control the quality of future human motions. 

\subsection{Problem Formulation}
An overview of the proposed DivDiff method is illustrated in Fig. 3.
The observed historical sequence is denoted as $\textbf{\emph{X}} =  \left\{\textbf{\emph{x}}_1, \textbf{\emph{x}}_2, \cdots, \textbf{\emph{x}}_T\right\}$ with length $T$, and the predicted human sequences is denoted as $\textbf{\emph{Y}} =  \left\{\textbf{\emph{Y}}_{T+1}, \textbf{\emph{Y}}_{T+2}, \cdots, \textbf{\emph{Y}}_{T+N} \right\}$ with length $N$, where $\textbf{\emph{x}}_t \in \mathbb{R}^{3\times J}$ is denoted by 3D coordinates at time $t$, $J$ is the number of human joints in a frame. Our goal is to predict diverse and realistic human motions through denoising process. 
Specifically, the forward process gradually introduces Gaussian noise to the input motions, ending with a whitened latent state. 
Conversely, the reverse process reconstructs the diverse future motions with a parameterized Markov chain. The DCT and transformer mechanisms are leveraged to encode the spatio-temporal information of human motions as a condition, so as to produce more smooth and realistic human motions. 
The DCT and Transformer operations extract both periodic temporal properties and spatial relations from the motion sequence, which is beneficial for obtaining continuous motions.
At the same time, the DRSF is introduced into the reverse process to enhance the diversity and quality of the predicted human motions.

\subsection{Preliminaries}
\textbf{Discrete Cosine Transform} 
The Discrete Cosine Transform (DCT) operation extracts both current and periodic temporal properties from the motion sequence, which is beneficial for obtaining continuous motions. The proposed method leverages DCT to capture the smoothness property of human motions. 

Given a human sequence $\textbf{\emph{X}}$, the proposed method projects the sequence into the DCT domain through the following operation: \\
\begin{equation}
   \mathbf{y}=\operatorname{DCT}(\mathbf{x}), 
\end{equation}
The DCT operation is an orthogonal transform, we can recover the motion sequence from the DCT domain via an inversed iDCT operation, which is formulated as follows:\\
\begin{equation}
    \mathbf{x}=\operatorname{iDCT}(\mathbf{y})
\end{equation}

\textbf{Spatial Transformer}
As shown in Fig. 3, the spatial transformer mainly focuses on the inter-dependencies among human joints within the same time step.
In the spatial transformer module process, given a human pose embedding at time $t$, $\boldsymbol{E}_t=[\boldsymbol{e}^{(1)}_{t},\dots, \boldsymbol{e}^{(a)}_{t}]^T $ and the weight matrices $\boldsymbol{W}^{(O)}$, the summary of the spatial joints $ \boldsymbol{\tilde{E}}_{t}$ is calculated by aggregating all the joint information at time $t$ using the multi-head mechanism. The formula for the calculation process of $ \boldsymbol{\tilde{E}}_{t} $ is defined as follows:
\begin{equation} \label{9}
	\begin{split}
		head_i &=Attention(\boldsymbol{Q}^{(i)}_t,\boldsymbol{K}^{(i)}_t, \boldsymbol{V}^{(i)}_t) \\
		 \tilde{\boldsymbol{E}}_t &=Concat(head_1,\dots,head_H)\boldsymbol{W}^{(O)}
	\end{split}
\end{equation}

For the spatial branch, the whole process operates on the joints at the same time-step.


\subsection{Motion Diffusion Process}
As illustrated in Fig. 3, the initial future motion $\textbf{\emph{Y}}^{0}$ will progressively diffuse into a whitened noise $\textbf{\emph{Y}}^{K}$, where $K$ is the number of diffusion steps. The diversity of the diffusion model relies on the diffusion process ($\textbf{\emph{Y}}^{0},\textbf{\emph{Y}}^{1}, \cdots, \textbf{\emph{Y}}^{K}$), which is defined as a fixed (without any trainable parameters) posterior distribution $q(\textbf{\emph{Y}}^{1:K}\mid\textbf{\emph{Y}}^{0})$. Formally, the posterior distribution of the diffusion process from $\textbf{\emph{Y}}^{0}$ to $\textbf{\emph{Y}}^{K}$ as:
\begin{equation}
  \begin{aligned}
  q\left(\mathbf{Y}^{1: K} \mid \mathbf{Y}^0\right)&=\prod_{k=1}^K q\left(\mathbf{Y}^k \mid \mathbf{Y}^{k-1}\right),\\
q\left(\mathbf{Y}^k \mid \mathbf{Y}^{k-1}\right)&=\mathcal{N}\left(\mathbf{Y}^k; \sqrt{1-\beta_k} \mathbf{Y}^{k-1}, \beta_k \mathbf{I}\right),  
  \end{aligned}
  \label{equation1}
\end{equation}
where $\beta_1, \beta_2, \cdots,\beta_k $ are pre-determined variance schedulers that control the injected noise in the diffusion process.  
Assuming $\alpha_k=1-\beta_k $ and $\Bar{\alpha_k}=\prod_{s=1}^k  \beta_s $, the forward process is computed for any step $k$ in a closed form as \cite{ho2020denoising}:
\begin{equation}
q\left(\mathbf{Y}^k \mid \mathbf{Y}^0\right) =\mathcal{N}\left(\mathbf{Y}^k ; \sqrt{\Bar{\alpha_k}} \mathbf{Y}^0,\left(1-\Bar{\alpha_k} \right) \mathbf{I}\right)
\end{equation}
This process indicates that $\textbf{\emph{Y}}^{0}$ will be converted into the isotropic Gaussian distribution when gradually adding new noises within $K$ steps. 


\subsection{The Conditional Reverse Process}
In the reverse process, we propose to encode the observed motion sequence $\textbf{\emph{X}}$ into state embedding $\mathcal{S}$ through DCT and transformer mechanisms, denoted by $f_\psi(\textbf{\emph{X}})$ with the parameter $\psi$, which serves as a condition $\mathcal{C}$ to guide the whole reverse process of DDPM. 
The human motion prediction process is formulated as a conditional reverse process ($\textbf{\emph{Y}}^{K}, \cdots, \textbf{\emph{Y}}^{1}, \textbf{\emph{Y}}^{0} $) with parameterized Gaussian transitions, starting from the white noise $\textbf{\emph{Y}}^{K} \sim p(\textbf{\emph{Y}}^{K}) $. 
Formally, the calculation process is defined as follows: 
\begin{equation}
\begin{gathered}
    p_\theta\left(\mathbf{Y}^{0: K-1} \mid \mathbf{Y}^K, \mathcal{C}\right) =\prod_{k=1}^K p_\theta\left(\mathbf{Y}^{k-1} \mid \mathbf{Y}^k, \mathcal{C}\right), \\
       p_\theta\left(\mathbf{Y}^{k-1} \mid \mathbf{Y}^k, \mathcal{C}\right) =\mathcal{N}\left(\mathbf{Y}^{k-1} ; \boldsymbol{\mu}_\theta\left(\mathbf{Y}^k, k, \mathcal{C}\right), \boldsymbol{\Sigma}_\theta\left(\mathbf{Y}^k, k\right)\right)
\end{gathered}  
\end{equation}

The distribution $p(\mathbf{Y}^K) $ is set as a standard Gaussian, $\mu_\theta(\mathbf{Y}^k, k )$ is the estimated mean parameterized by $\theta$, and $\Sigma_\theta(\mathbf{Y}^k, k)$ is empirically set to $\beta_k \mathbf{I}$. 
In total, given a conditional $\mathcal{C}$ and chaotic states $\mathbf{Y}^K$ from $p(\mathbf{Y}^K)$, we then gradually denoise through the reverse Markov kernels 
$p_\theta(\mathbf{Y}^{k-1} \mid \mathbf{Y}^k, \mathcal{C})$ to predict diverse human motions. 
However, after extensive experiments, we have found that there is a trade-off between the motion plausible and diversity when varying the number of denoising steps. Existing studies published to date on human motion prediction have rarely been considered to address this problem. Therefore, we introduce the DRSF function into the late stage of the reverse process.


\begin{algorithm} [tb]
  \caption{The Diversified Reinforcement Sampling Algorithm.}
  \label{alg1}
  \begin{algorithmic}[1]
      \Require
        The state embedding $\{\mathbf{x}^{(i)}, \boldsymbol{\mathcal{\emph{S}}}^{(i)}\}_{i=1}^N$, DDPM $D_{\theta}(\epsilon,\mathcal{\emph{S}},t)$, sample time $t$, target time $s$
      \Ensure
        The diversified sampling function $g_{\gamma}(\mathcal{\emph{S}})$
      \State Initialize $\gamma$ randomly
      \If{sample time $t$ \textgreater time $s$}
      \While {not converged }
         \For{each $\psi^{i}$}  
            \State Generate noises $\epsilon_\theta=\left\{\epsilon_1, \ldots, \epsilon_N\right\}$ with $g_{\gamma}(\mathcal{\emph{S}})$ 
            \State Generate the predicted future motions $\mathbf{\emph{Y}}={\mathbf{\emph{Y}}_{T+1},...,\mathbf{\emph{Y}}_{T+N}}$ with the DDPM $D_{\theta}(\epsilon,\mathcal{\emph{S}},t)$ 
            \State Compute the similarity matrix $\mathcal{S}$ and quality vector $\textbf{r}$ with Eq. (7) and Eq. (8)
            \State Compute the DPP kernel $\mathbf{L}(\gamma) = Diag(\mathbf{r})\cdot \mathbf{S}\cdot Diag(\mathbf{r})$
            \State Calculate the final loss $\emph{L}_{diverse}(\gamma)$
            \State Update $\gamma$ with gradient $\nabla$ $\emph{L}_{diverse}(\gamma)$
         \EndFor 
        \EndWhile
        \Else
        \State Generate noises $\epsilon_\theta=\left\{\epsilon_1, \ldots, \epsilon_N\right\}$  with DDPM $f_{\theta}(\epsilon,\mathcal{\emph{S}},t)$
        \State Generate the predicted future motions $\mathbf{\emph{Y}}={\mathbf{\emph{Y}}_{T+1},...,\mathbf{\emph{Y}}_{T+N}}$ with the DDPM $D_{\theta}(\epsilon,\mathcal{\emph{S}},t)$ 
        \EndIf
  \end{algorithmic}
\end{algorithm}

\subsection{The Diversified Reinforcement Sampling Strategy}
As previously mentioned, introducing DDPM into human motion prediction cannot guarantee the high diversity and fidelity of predicted human motions. 
The noise parameter $\beta$ fluctuates significantly in the late stage of the forward process. This causes significant disturbances in the reverse process, making it difficult to control the predicted human motions.
In this work, we propose a new strategy (DRSF) for diversified reinforcement sampling that utilizes human skeleton information as prior to predict random noise during the later stages of the reverse process.


DRSF involves the determinantal point process function (DPP) to improve the diversity of predictions. 
DPP is an efficient probabilistic function to enhance diversity by capturing negative correlations between samples, which has been successfully applied in many tasks, e.g., document summarization \cite{gong2014diverse}, object detection \cite{azadi2017learning} and diverse trajectory forecasting \cite{yuan2019diverse}. Our method aims to test the influence of DPP on DDPM for future human motion prediction.
DPP models the probability of encompassing a sample while reducing the probability of including similar samples, making it an appropriate tool to capture diversity in a set. In particular, the DRSF is implemented by 3-layer $\gamma$-parameterized graph convolutional networks, denoted by $g_{\gamma}(\mathcal{\emph{S}})$. 
In the late stage of the reverse process, the $g_{\gamma}(\mathcal{\emph{S}})$ takes the state embedding information $\mathcal{\emph{S}}$ as the input, which contains the human skeleton constraints to output random noises. 
Then, DDPM samples according to the random noises and decodes them into human motions, i.e., $g_{\gamma}(\mathcal{\emph{S}}) \to \epsilon$.
To optimize the DRSF, a diversity loss based on DPP is introduced into this work. The detailed procedure is illustrated in Algorithm 1. The following describes how the DPP kernel $\mathbf{L}$ constructs a loss function so as to optimize the DRSF.

DPP establishes the probability associated with selecting a random subset from Gaussian distributions, which involves the \textbf{Similarity} matrix $\mathbf{S}$ between two motions, and the \textbf{Quality} vector $\mathbf{r}$ of each generated human motion, i.e., $\mathbf{L}(r) = Diag(\mathbf{r})\cdot \mathbf{S}\cdot Diag(\mathbf{r})$. In this work, DPP kernel $\mathbf{L}(\gamma)$ is formulated based on $\gamma$ as it is defined on the ground truth $\mathbf{\emph{Y}}$ decoded by the DRSF $g_{\gamma}(\mathcal{\emph{S}})$.

\textbf{Similarity}: Regarding similarity, the Euclidean distance can serve as a simple yet efficient metric between human motions, then the similarity $\mathbf{S}_{ij}$ between two motions $\mathbf{y}_i$ and $\mathbf{y}_j $ are defined as follows:
\begin{equation}\label{eq:similarity}
\mathbf{S}_{ij}=\exp (-k\left\|\mathbf{y}_i-\mathbf{y}_j\right\|^2),
\end{equation}
where $k$ is the scaling vector. This operation assures $0\leq\mathbf{S}_{ij}\leq1$ and $\mathbf{S}_{ii}=1$, and $\mathbf{S}$ a is positive matrix.

\textbf{Quality}: Different from the similarity metric, the quality metric is measured as follows:
\begin{equation}\label{eq:quality}
r_i= \begin{cases}\omega, & \text { if }\left\|\mathbf{m}_i\right\| \leq R \\ \omega \exp \left(-\mathbf{m}_i^T \mathbf{m}_i+R^2\right), & \text { otherwise }\end{cases},
\end{equation}
where $R$ is the radius of a sphere $\phi$ that encompasses the majority of samples from the Gaussian prior distribution, and $\mathbf{m}_i$ is the feature representation of human motion.
In this way, samples within the boundary defined by $\phi$ are treated equally, regardless of whether they belong to major or minor patterns. Samples located far away from the data manifold receive a heavy penalty as they fall outside of $\phi$.
$\omega$ denotes the base quality, which controls the item selection in DPP. A larger value of $\omega$ leads the DPP to choose a greater number of items from a whole set.

We introduce the following loss function $L_{diverse}(\gamma)$ to optimize the DRSF $g_{\gamma}(\mathcal{\emph{S}})$.
The diversity of $\mathbf{\emph{Y}}$ is assessed using the expected cardinality of the DPP, defined as $\mathbb{E}_{\boldsymbol{Y} \sim \mathcal{P}_{\mathbf{L}(\gamma)}}[|\boldsymbol{Y}|]$. Essentially, a randomly drawn subset following the DPP is inclined to include a larger number of diverse items from $\mathbf{\emph{Y}}$, given that the DPP discourages the selection of similar items. The expected cardinality offers the advantage of being well-defined even when the ground set $\mathbf{\emph{Y}}$ contains duplicate items. Formally, the expected cardinality is defined as follows:
\begin{equation} \mathbb{E}[|\boldsymbol{Y}|]=\sum_{n=1}^N \frac{\lambda_n}{\lambda_n+1}=\operatorname{tr}\left(\mathbf{I}-(\mathbf{L}(\gamma)+\mathbf{I})^{-1}\right), \end{equation}
where $\lambda_n$ is the $n$-th eigenvalue of $\mathbf{L}$ and $\mathbf{I}$ denotes the identity matrix. The diversity loss is defined as follows:
\begin{equation} 
\begin{aligned}
    L_{\text {diverse }}(\gamma)&=-\operatorname{tr}\left(\mathbf{I}-(\mathbf{L}(\gamma)+\mathbf{I})^{-1}\right)  ,\\
\end{aligned}
\end{equation}

The  ${L}_{\text {diverse }}(\gamma)$ is employed to optimize the DRSF.

\begin{table*}[]
\label{table1}
\caption{The comparison results between the proposed DivDiff and state-of-the-art methods on Human3.6M and HumanEva-I datasets. The best results are in bold. Lower is better for all metrics except the APD metric.}
\resizebox{\linewidth}{!}{
\begin{tabular}{c|c|ccccc|ccccc}
\hline
\multirow{2}{*}{Type}          & \multirow{2}{*}{Method} & \multicolumn{5}{c|}{Human3.6M}                                                      & \multicolumn{5}{c}{HumanEva-I}                                                      \\ \cline{3-12} 
     &     & APD $\uparrow$    & ADE $\downarrow$  & FDE $\downarrow$    & MMADE $\downarrow$         & MMFDE $\downarrow$    
           & APD $\uparrow$    & ADE $\downarrow$  & FDE $\downarrow$    & MMADE $\downarrow$         & MMFDE $\downarrow$         \\ \hline
\multirow{2}{*}{Deterministic} & LTD(ICCV'19)            & 0.000           & 0.516          & 0.756          & 0.627          & 0.795          & 0.000           & 0.415          & 0.555          & 0.509          & 0.613          \\
                               & MSR(ICCV'21)            & 0.000           & 0.508          & 0.742          & 0.621          & 0.791          & 0.000           & 0.371          & 0.493          & 0.472          & 0.548          \\ 
                               & acLSTM (ICLR’18)      & 0.000           & 0.789    &1.126          & 0.849          & 1.139             & 0.000         & 0.429          & 0.541             & 0.530          & 0.608          \\
                               \hline
\multirow{9}{*}{Stochastic}    & Pose-Konws(ICCV'17)     & 6.723           & 0.461          & 0.560          & 0.522          & 0.569          & 2.308           & 0.269          & 0.296          & 0.384          & 0.375          \\
                               & MT-VAE(ECCV'18)         & 0.403           & 0.457          & 0.595          & 0.716          & 0.883          & 0.021           & 0.345          & 0.403          & 0.518          & 0.577          \\
                               & GMVAE(arKiv'16)         & 6.769           & 0.461          & 0.555          & 0.524          & 0.566          & 2.443           & 0.305          & 0.345          & 0.408          & 0.410          \\ \cline{2-12} 
                               & DLow(ECCV'20)           & 11.741          & 0.425          & 0.518          & 0.495          & 0.531          & 4.855           & 0.251          & 0.268          & 0.362          & 0.339          \\
                               & HP-GAN(CVPRW'18)        & 7.214           & 0.858          & 0.867          & 0.847          & 0.858          & 1.139           & 0.772          & 0.749          & 0.776          & 0.769          \\
                               & GSF(ICLR'19)            & 9.330           & 0.493          & 0.592          & 0.550          & 0.599          & 4.538           & 0.273          & 0.290          & 0.364          & 0.340          \\
                               & MOJO(CVPR'21)           & 12.579          & 0.412          & 0.514          & 0.497          & 0.538          & 4.181           & 0.234          & 0.244          & 0.369          & 0.347          \\
                               & DivSamp(ACM MM'22)      & 15.310          & 0.370          & 0.485          & 0.475          & 0.516          & \textbf{24.724} & 0.564          & 0.647          & 0.623          & 0.667          \\
                               & BeLFusion(arXiv'22)     & 7.602           & 0.372          & \textbf{0.474} & 0.473          & 0.507          & 9.376           & 0.513          & 0.560          & 0.569          & 0.585          \\ 
                               & MotionDiff(AAAI'23)     & 15.353           & 0.411          & 0.509 & 0.508          & 0.536          & 5.931           & 0.232          & 0.236          & 0.352          & 0.320          \\
                               \cline{2-12} 
                               & DivDiff(Ours)           & \textbf{15.602} & \textbf{0.360} & 0.503          & \textbf{0.443} & \textbf{0.403} & 10.430          & \textbf{0.210} & \textbf{0.216} & \textbf{0.314} & \textbf{0.209} \\ \hline
\end{tabular} }
\end{table*}

In addition, according to \cite{ho2020denoising}, the objective training function of the early stage of the reverse process in DDPM is summarized as a simplified loss function:
\begin{equation}
L(\psi, \theta)=\mathbb{E}_{k, \mathbf{\emph{Y}}^0, \epsilon}\left[\left\|\epsilon-\epsilon_\theta\left(\mathbf{Y}^k, k, f_\psi(\mathcal{S})\right)\right\|^2\right],
\end{equation}
where $k$ is the diffusion step, $f_\psi(\mathcal{S})$ denotes the encoded state embedding $\mathcal{S}$ with the parameter $\psi$ and $\epsilon_\theta$ is the predicted noise in each diffusion step.

In brief, the overall loss in the proposed method is a weighted sum of the aforementioned individual loss functions. The formula for the global loss, denoted as $L_{total}$, is articulated as follows:
\begin{equation}
     L_{total}=\alpha_1L_{diverse}(\gamma)+\alpha_2L(\psi, \theta)+ \alpha_3\min _{i}\left\|\mathbf{Y}_{i}-\mathbf{X}\right\|^{2},
\end{equation}

$L_{total}$ is utilized to balance the diversity and the accuracy of the results. The third loss function is used to ensure that the distance of predicted motions is not too far away from the observed motion, thus guaranteeing realism and accuracy. The parameters $\alpha_1$, $\alpha_2$, and $\alpha_3$ represent the coefficients for different losses. These values are determined empirically in our experiments. Specifically, we set $\alpha_1=0.4$, $\alpha_2=0.3$, and $\alpha_3=0.3$ as the loss coefficients.

\textbf{Sampling}: Given a learned reverse diffusion network $\epsilon_\theta$, an observed sequence $\mathbf{\emph{X}}$ and its corresponding encoder $f_\psi(\mathcal{\emph{S}})$, we first sample chaotic states $\mathcal{\emph{Y}}^K$ from $\mathcal{N}(0,\mathbf{I})$.
Subsequently, we progressively predict realistic human motions from $\mathcal{\emph{Y}}^K$ to
$\mathcal{\emph{Y}}^0$ by the following formula:

\begin{equation}
   \mathbf{\emph{Y}}^{k-1} =\frac{1}{\sqrt{\alpha_k}}\left(\mathbf{\emph{Y}}^k-\frac{\beta_k}{\sqrt{1-\bar{\alpha}_k}} \epsilon_\theta\left(\mathbf{\emph{Y}}^k, k, f_\psi(\mathcal{\emph{X}})\right)\right)
    +\sqrt{\beta_k} \mathbf{z},
\end{equation} 
where $\mathbf{z}$ is a random variable from standard Gaussian distribution.


\section{Experimental Design}
In this section, the experimental design will be described, encompassing benchmark datasets, parameter settings, evaluation metrics, and baseline methods.

\subsection{Datasets}
To keep consistent with the latest methods \cite{yuan2022physdiff,ma2022multi}, we evaluate the proposed method on two widely-used datasets, Human3.6M (H3.6M) \cite{ionescu2013human3} and HumanEva-I \cite{sigal2010humaneva}. The details are as follows.

\textbf{Human3.6M}: Human3.6M is a widely utilized dataset for 3D human motion prediction, featuring 15 motions performed by 7 actors. We preprocess the dataset through a down-sampling operation to enhance the training efficiency, reducing the frame rate (in frames per second, fps) from 50 to 25. This down-sampling operation aims to make the inherent motion patterns clearer. The proposed method is trained on data from 5 subjects (S1, S5, S6, S7, S8), while subjects S9 and S11 are reserved for testing and validation. In this study, we input 25 frames, equivalent to 0.5 seconds at 50fps, to predict 100 frames.

\textbf{HumanEva-I}: HumanEva-I consists of 3 subjects, which are performed in 5 action categories, depicted by video captured at 60 Hz. A person is represented by a skeleton with 15 joints. We adopt the official train/test split \cite{yuan2020dlow} and also remove the global translation. The proposed method predicts 60 future poses (1s, 60fps) given 15 past frames. 

\subsection{Parameter Settings}
For the diffusion network, the proposed DivDiff method utilizes an encoder to upsample the hidden dimension of 3D coordinates of human joints from 3d to 512d, which consists of DCT and transformer mechanisms. Then, the DivDiff employs a decoder that has the same structure to decode the dimension to 3d. The DSRF is implemented by a 3-layer GCN. 
Our method predicts 50 diverse future motions ($N$ = 50) given a past motion. In contrast with other methods, the number of steps $K$ in the diffusion process is set to 200.
The variance schedules is set to be $\beta_1=$0.0001 and $\beta_K=$0.05, where $\beta_K$ are linearly interpolated (1 $\textless$ $k$ $\textless$ $K$). 
The proposed DivDiff method is implemented using the PyTorch framework in Python 3.6. To ensure the convergence of the proposed method, we employ the Adam optimizer for model training. The learning rate is initially set to $10^{-4}$ and undergoes a decay of 0.98 every 10 epochs. Training for the proposed method spans 500 epochs, utilizing a batch size of 64 for both training and evaluation. All experiments are conducted on the Nvidia Tesla A100 GPU.

\subsection{Evaluation Metrics}
In order to contrast with other methods, we employ the following metrics to measure the diversity and accuracy of the proposed method.
(1) APD: Average Pairwise Distance between all pairs of motion samples defined as $\frac{1}{N(N-1)} \sum_{i=1}^N \sum_{j \neq i}^N\left\|\mathbf{Y}_i-\mathbf{Y}_j\right\|_2$.
(2) ADE: Average Displacement Error over the whole sequence between the ground truth and the closest generated motion defined as $\frac{1}{f} min_i \left\|\mathbf{Y}_i-\mathbf{X}\right\|_2 $.
(3) FDE: Final Displacement Error between the last frame of the ground truth and the closest motion’s last frame defined as $ min_i \left\|\mathbf{Y}_i[f] -\mathbf{X}[f]\right\|_2 $.
(4) MMADE: the multi-modal version of ADE; (5) MMFDE: the multi-modal version of FDE. Note that ADE is utilized to measure the diversity while others are utilized to measure the accuracy of the proposed method. 

\subsection{Baseline Methods}
Two types of baselines (deterministic and stochastic motion prediction methods) are leveraged in this work to evaluate the performance of the proposed method from the aspect of accuracy and diversity.
(1) Deterministic motion prediction methods, including LTD \cite{mao2019learning}, MSR \cite{dang2021msr} and acLSTM \cite{li2018structure}.
(2) Stochastic motion prediction methods, including Pose-Konws \cite{walker2017pose}, MT-VAE \cite{yan2018mt}, GMVAE \cite{dilokthanakul2016deep}, DLow \cite{yuan2020dlow}, HP-GAN \cite{barsoum2018hp}, DSF \cite{yuan2019diverse}, MOJO \cite{zhang2021we}, DivSamp \cite{dang2022diverse}, BeLFusion \cite{barquero2022belfusion} and MotionDiff \cite{wei2022human}.

\begin{figure*}[htb]
 \centering 
 \includegraphics[width=6.2in]{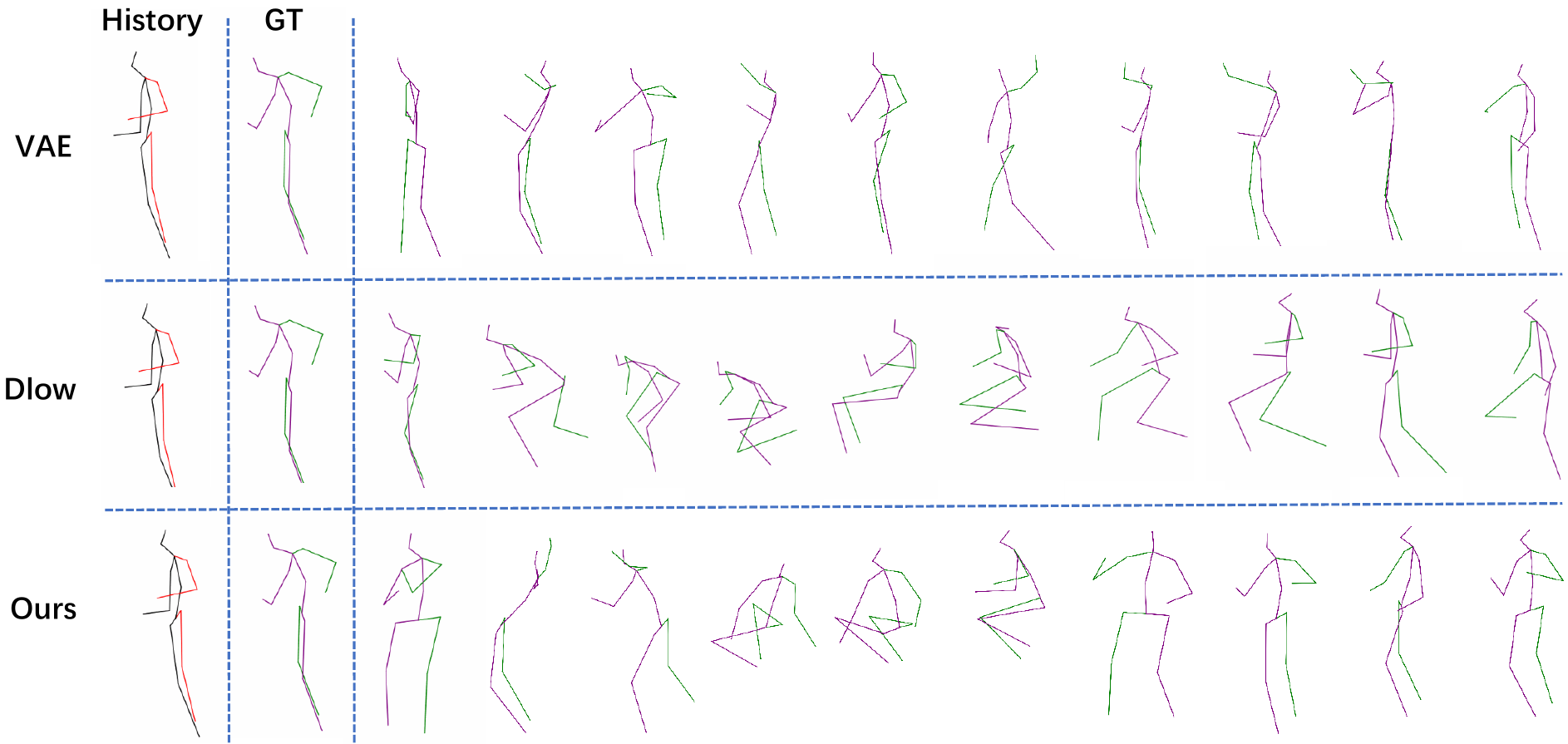}
 \caption{Qualitative comparison between other methods and the proposed DivDiff method. Given the observed motion sequence, the figure shows the end poses of ten future predictions. The proposed DivDiff method yields human motions with more diverse and realistic.}
 \label{fig:4}
\end{figure*}

\begin{figure}[htb]
\resizebox{\linewidth}{!}{
 \centering 
 \includegraphics[width=6.9in]{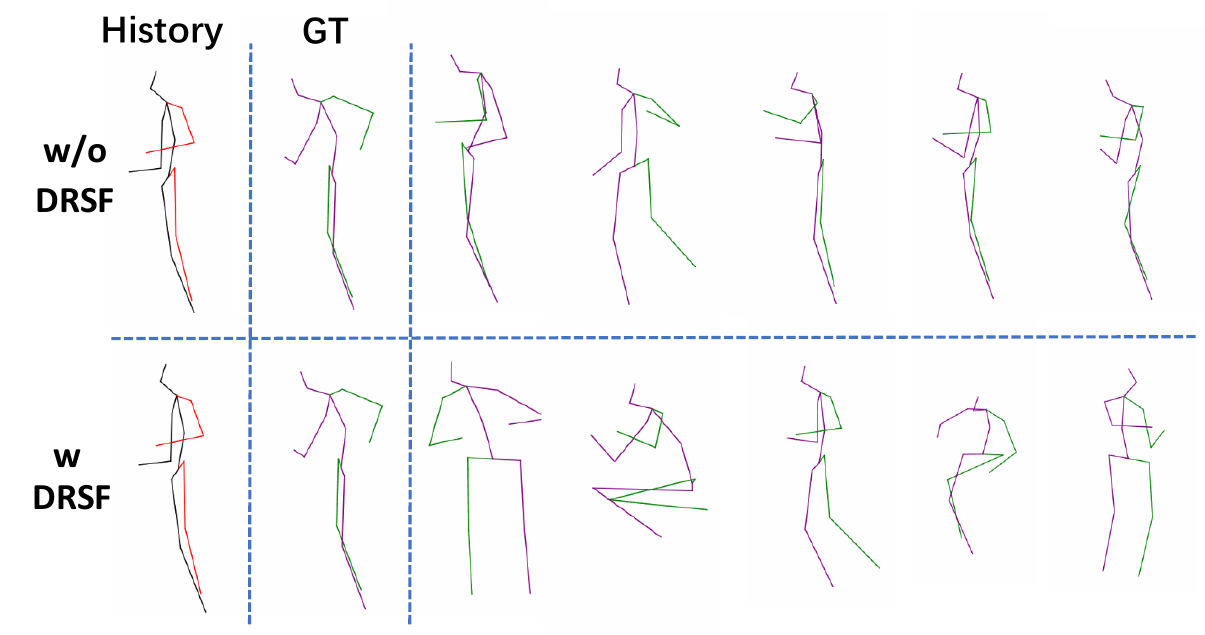} }
 \caption{Comparison results of the proposed DivDiff method with (bottom) and without (top) DRSF. The results obtained from the proposed DivDiff method without DRSF exhibit lower diversity when compared to those obtained from the proposed DivDiff method with DRSF.}
 \label{fig:5}
\end{figure}

\section{Results and Analysis}
The section endeavors to provide a comprehensive analysis of the experimental results for accuracy and diversity, including the quantitative comparison results between the state-of-the-art methods and the proposed method, ablation analysis and qualitative analysis.

\subsection{Comparision to Existing Methods}
Table 1 summarizes the comparison prediction results under multiple baseline methods on Human3.6M and HumanEva-I datasets, DivDiff achieves state-of-the-art performance in accuracy and diversity metrics for both datasets. From the empirical evidence, it is observed that the proposed DivDiff method consistently outperforms all the baselines based on all the evaluation metrics.
For the deterministic approaches, the diversity results are 0.000 since these approaches only predict one output. The prediction accuracy is also inferior to stochastic methods. We speculate that deterministic prediction models tend to predict an average mode as the final result, which leads to higher prediction errors.
For the stochastic methods, it can be observed that the DivDiff method outperforms other methods by a large margin in terms of diversity and accuracy, including the DDPM-based method under the same denoising steps. This pronounced discrepancy in performance can likely be attributed to a couple of key factors. The foremost factor is the strategic utilization of the DCT and transformer mechanisms as conditional elements for forecasting future motions. This incorporation serves to enhance the overall smoothness and accuracy of the results. Moreover, the DDPM, bolstered by its ingeniously designed DRSF, plays a pivotal role in augmenting the diversity within the predicted human motions. This synergy between methodology and function imparts a distinct advantage in predicting a wide spectrum of human motions.

We further demonstrate the effectiveness of the proposed method through the qualitative results. As illustrated in Fig. 4, we provide a visual comparison between our DivDiff method and several contemporary state-of-the-art approaches, including those discussed in \cite{yuan2020dlow} and \cite{barquero2022belfusion}. The comparison results readily reveal the distinctive attributes of our approach.
Evidently, our method stands out by predicting human motions that exhibit a remarkable degree of diversity and realism, far surpassing the performance of the competing baselines. This compelling visual evidence clearly demonstrates the superiority of our method.

\begin{figure}[htb]
\resizebox{\linewidth}{!}{
 \centering 
 \includegraphics[width=4.2in]{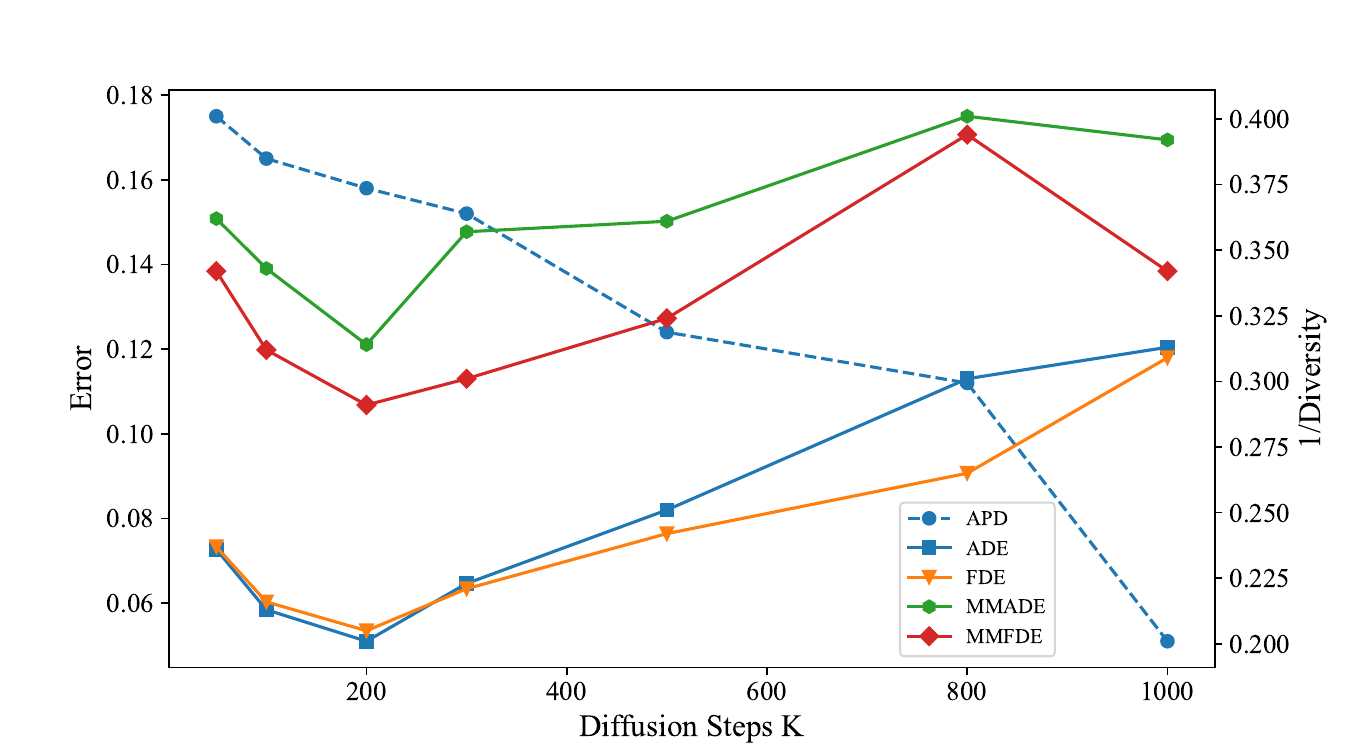} }
 \caption{The diversity and prediction errors of the proposed DivDiff method with different diffusion steps $K$.}
 \label{fig:6}
\end{figure}

\begin{table*}[!ht]
\label{tab2}
\caption{Comparison results on two datasets in terms of different components of the proposed DivDiff .}
\resizebox{\linewidth}{!}{
\begin{tabular}{ccc|ccccc|ccccc}
\hline
\multirow{2}{*}{DCT} & \multirow{2}{*}{Transformer} &\multirow{2}{*}{DRSF}  & \multicolumn{5}{c|}{Human3.6M}                                                      & \multicolumn{5}{c}{HumanEva-I}                                                     \\ \cline{4-13} 
                     &           &     & APD $\uparrow $     & ADE $\downarrow$           & FDE $\downarrow$     & MMADE $\downarrow$   & MMFDE $\downarrow$     & APD $\uparrow$       & ADE $\downarrow$      & FDE $\downarrow$      & MMADE  $\downarrow$        & MMFDE  $\downarrow$     \\ \hline
             \usym{2713}      &  \ding{55}   &  \ding{55}       & 12.613          & 0.413          & 0.526          & 0.526          & 0.596          & 8.142           & 0.306          & 0.291          & 0.372          & 0.310         \\
             \usym{2713}  &   \usym{2713}  &   \ding{55}        & 12.720          & 0.401          & 0.514          & 0.514          & 0.531          & 8.351           & 0.268          & 0.276          & 0.362          & 0.247         \\
            \usym{2713}   &  \usym{2713}          &   \usym{2713}                  & \textbf{15.602} & \textbf{0.360} & \textbf{0.503} & \textbf{0.443} & \textbf{0.403} & \textbf{10.430} & \textbf{0.210} & \textbf{0.216} & \textbf{0.314} & \textbf{0.209} \\ \hline
\end{tabular}}
\end{table*}

\subsection{Ablation Studies}
In this subsection, ablation studies are conducted to examine the effectiveness of different components of the proposed DivDiff method quantitatively. As described in Table 2, introducing the DCT into this work did not enhance the diversity performance dramatically, but improved the accuracy of the results. This proves that the introduced DCT method is beneficial to this task.
Notably, incorporating the proposed DRSF significantly promotes the performance both in diversity and accuracy, which can capture more modes of human motions. Fig. 5 shows the effectiveness of our choice by qualitative results. It can be seen that the diversity of the motions improved through the DRSF (below).

To evaluate the influence of the different number of diffusion steps $K$ in the proposed method, we provide an analysis of the results between diversity and prediction error on the Human3.6M dataset. As shown in Fig. 6, when $K$ is small, the predicted human motions are more diverse (measured by APD) but usually lead to low accuracy (except when $K$ approaches 200). 
The other metrics (ADE, FDE, MMADE, and MMFDE) all worsen as the diffusion steps increase. This is because more diffusion steps lead to the predicted close to the ground truth, while fewer diffusion steps can generate more diverse, yet implausible samples.
Compared to other methods where the error gradually increases when $K$ approaches 100 (the predicted motions are far from the
ground truth, highlighted by red boxes), our method can increase up to nearly 200 diffusion steps. 
In addition, Fig. 7 shows the change in predicted results with the increase of denoising steps $K$. It can be observed that the denoising step of 200 produced more diverse human motions compared to setting it to 100. Although a denoising step of 500 can enhance the slight accuracy, the diversity of the predicted motions declines dramatically, which also validates the effectiveness of the proposed method.

In addition to the aforementioned results, the experiments have also conducted a visual analysis of the training process. As illustrated in Fig. 8, comparing the training curves associated with and without the utilization of the DRSF function. This visualization offers valuable insights into the effectiveness of the proposed DRSF function.
Specifically, the blue curve represents the training process conducted without the incorporation of the DRSF function, while the green curve corresponds to the proposed DivDiff method in conjunction with the DRSF function. It is clearly evident that the integration of the proposed DRSF function results in a significant acceleration of the convergence speed, consequently enhancing the overall performance of our method. The comprehensive experiments demonstrated that our method achieves better performance with the proposed DRSF function.

The comprehensive experiments demonstrate that our method consistently outperforms the state-of-the-art methods. These findings serve to underscore the pivotal role of the proposed DRSF function, particularly in terms of the diversity and reality of the predicted human motions.

\begin{figure}[htb]
 \centering 
 \includegraphics[width=3.7in]{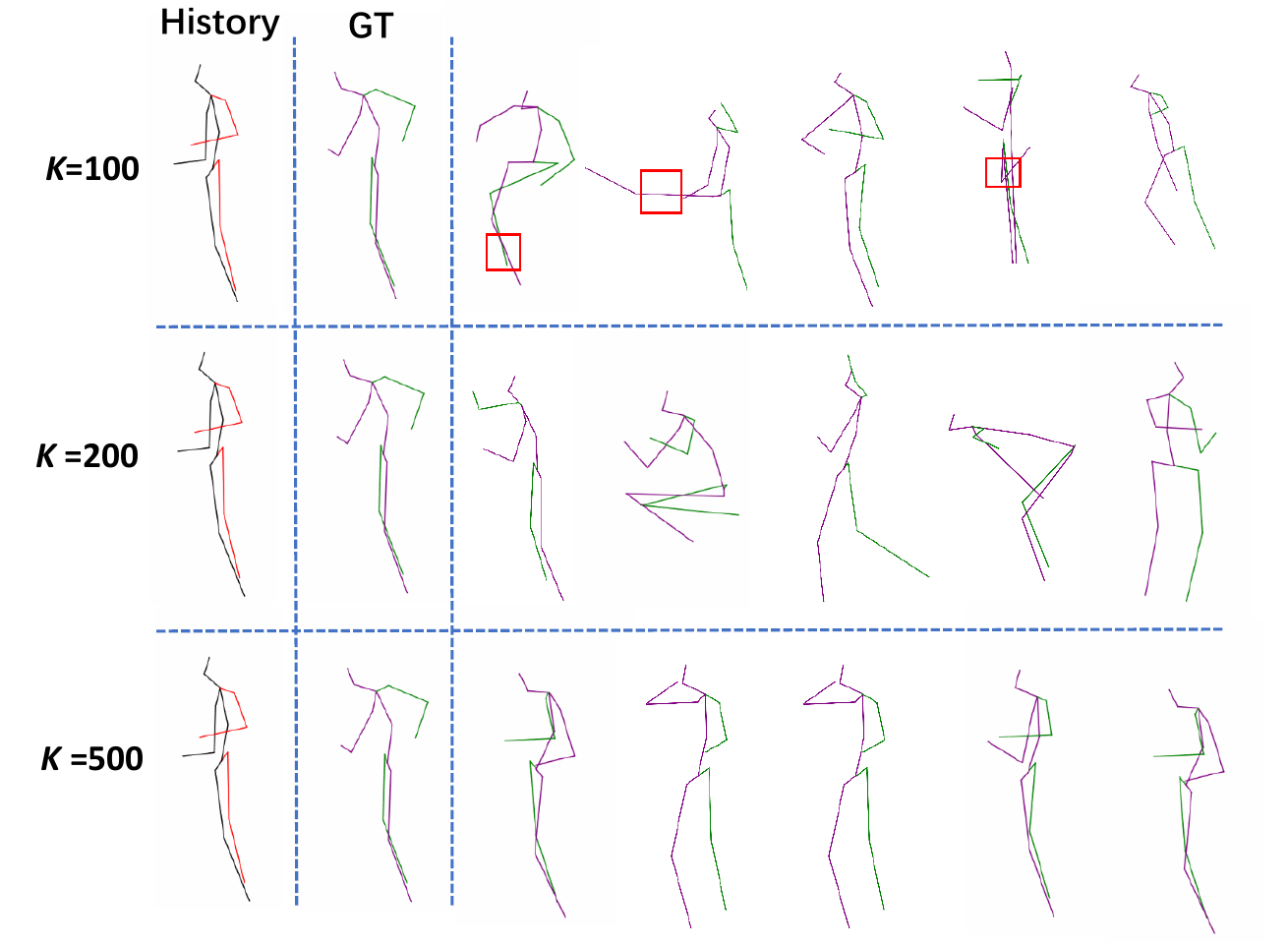}
 \caption{Visualization of predicted human motion sequences with different denoising steps.}
 \label{fig:7}
\end{figure}

\begin{figure}[htb]
 \centering 
 \includegraphics[width=3.7in]{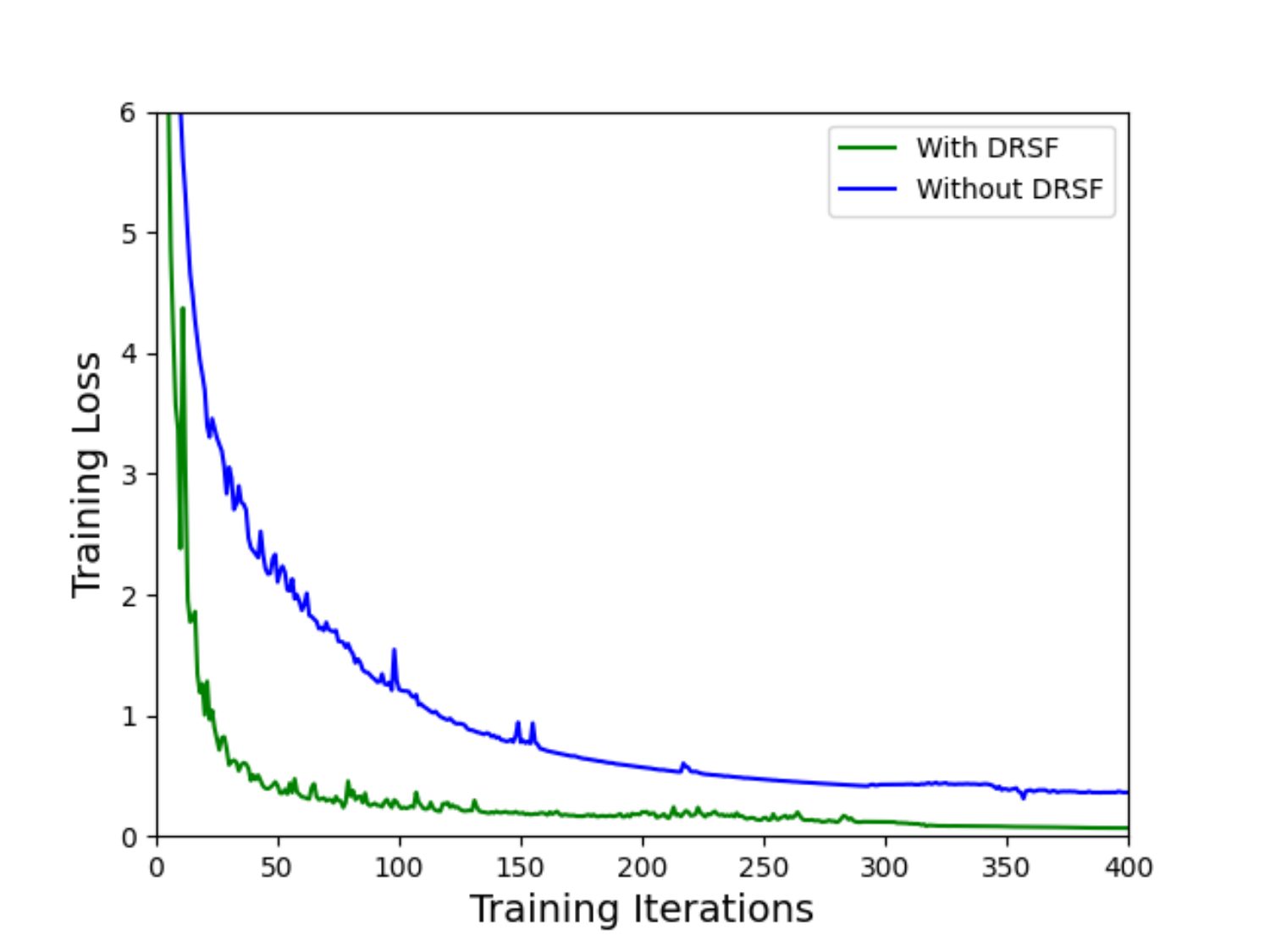}
 \caption{The training process involves comparing the curves of the proposed method with and without the introduced DRSF function.}
 \label{fig:8}
\end{figure}

\section{Conclusion}
The goal of this paper is to predict more diverse and realistic human motions based on the diffusion method. This has been achieved by designing a new method (called DivDiff), which combines the inherent relationships of human skeletons with a conditional DDPM to balance motion diversity and quality in the predicted human motions.
Specifically, the observed sequence is encoded by the DCT and transformer and utilized as a condition to instruct the reverse diffusion process. Moreover, the proposed DivDiff method utilizes the designed DRSF that leverages graph neural networks to add more human skeleton constraints on the predicted human motions. The DRSF serves as a prior to produce random noises, thereby alleviating significant disturbances of the reverse process. 

In the experiments, the proposed DivDiff method was compared with two types of human motion prediction methods (i.e., deterministic and stochastic motion prediction methods). Extensive experiments demonstrate that the proposed DivDiff method achieved superior performance in almost all cases.

In addition, considering that our method can alleviate the significant disturbance in the late stage of the reverse process in DDPM, the proposed DivDiff method also can be used for many generative-related tasks, such as human motion estimation and image synthesis. 



\bibliographystyle{unsrt}
\bibliography{main}

\vfill

\end{document}